\documentclass[10pt,twocolumn,letterpaper]{article}

\usepackage[pagenumbers]{cvpr} 

\usepackage{graphicx}
\usepackage{amsmath}
\usepackage{amssymb}
\usepackage{booktabs}
\usepackage{multirow}
\usepackage{bbding}
\usepackage{pifont} 
\usepackage{flushend}
\usepackage[table]{xcolor}
\usepackage{placeins}
\usepackage{bbold}
\usepackage[pagebackref,breaklinks,colorlinks]{hyperref}
\usepackage[capitalize]{cleveref}
\usepackage[bottom]{footmisc} 
\usepackage{enumitem}

\definecolor{lightred}{rgb}{1, 0.5, 0.5}
\definecolor{lightblue}{rgb}{0.5, 0.6, 1}
\crefname{section}{Sec.}{Secs.}
\Crefname{section}{Section}{Sections}
\Crefname{table}{Table}{Tables}
\crefname{table}{Tab.}{Tabs.}

\newcommand{{\engine}}{Traffic Manager}
\newcommand{\dreamer}{World Dreamer}
\newcommand{\arena}{\textsc{DriveArena}}
\newcommand{\yes}{\textcolor{lightblue}{\ding{52}}}
\newcommand{\no}{\textcolor{lightred}{\ding{55}}}

\begin{document}

\title{\raisebox{-1.0ex}{\includegraphics[height=2.0\baselineskip]{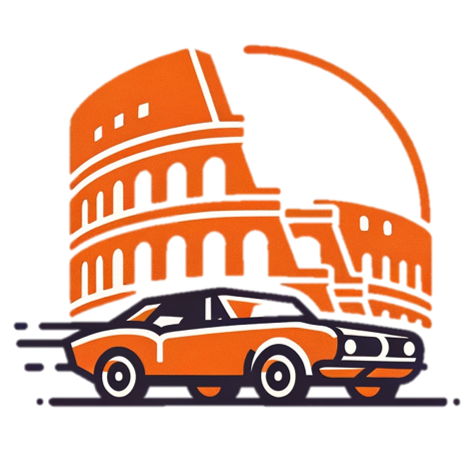}} \huge {\arena}: A Closed-loop Generative Simulation Platform for Autonomous Driving}

\author{\textbf{Xuemeng Yang$^{1, *}$ \quad Licheng Wen$^{1, *}$  \quad Yukai Ma$^{2, 1, *}$ \quad Jianbiao Mei$^{2, 1, *}$ \quad Xin Li$^{3,5, *, \dagger}$} 
\vspace{7pt} \\ 
\textbf{\quad Tiantian Wei$^{1, 4, *}$ \quad Wenjie Lei$^{2, \dagger}$ \quad Daocheng Fu$^{1}$ \quad Pinlong Cai$^{1}$ \quad Min Dou$^{1}$}
\vspace{7pt} \\
\textbf{\quad Botian Shi$^{1, \text{\Envelope}}$ \quad Liang He$^{5}$ \quad Yong Liu$^{2, \text{\Envelope}}$ \quad Yu Qiao$^1$} 
\vspace{7pt}
\\
$^{1}$ Shanghai Artificial Intelligence Laboratory \quad
$^{2}$ Zhejiang University \quad
$^{3}$ Shanghai Jiao Tong University 
\vspace{7pt}
\\
$^{4}$ Technical University of Munich \quad
$^{5}$ East China Normal University
\vspace{7pt}
\\
Project Page:~\url{https://pjlab-adg.github.io/DriveArena/}
}

\maketitle
\renewcommand{\thefootnote}{\relax} 
\setlength{\footnotemargin}{0pt} 
\setlength{\footnotesep}{1pt} 
\footnotetext{\noindent ${*}$ Equal contribution,~~ ${\text{\Envelope}}$ Corresponding author \\
${\dagger}$ Work performed during internships at Shanghai AI Laboratory}

\begin{abstract}
\normalsize
This paper presented {\arena}, the first high-fidelity closed-loop simulation system designed for driving agents navigating in real scenarios. {\arena} features a flexible, modular architecture, allowing for the seamless interchange of its core components: {\engine}, a traffic simulator capable of generating realistic traffic flow on any worldwide street map, and {\dreamer}, a high-fidelity conditional generative model with infinite autoregression. This powerful synergy empowers any driving agent capable of processing real-world images to navigate in {\arena}'s simulated environment. The agent perceives its surroundings through images generated by {\dreamer} and output trajectories. These trajectories are fed into {\engine}, achieving realistic interactions with other vehicles and producing a new scene layout. Finally, the latest scene layout is relayed back into {\dreamer}, perpetuating the simulation cycle. This iterative process fosters closed-loop exploration within a highly realistic environment, providing a valuable platform for developing and evaluating driving agents across diverse and challenging scenarios. {\arena} signifies a substantial leap forward in leveraging generative image data for the driving simulation platform, opening insights for closed-loop autonomous driving.

Code will be available soon on GitHub:~\url {https://github.com/PJLab-ADG/DriveArena}
\end{abstract}

\begin{figure}[t]
    \centering
    \includegraphics[width=1.1\linewidth]{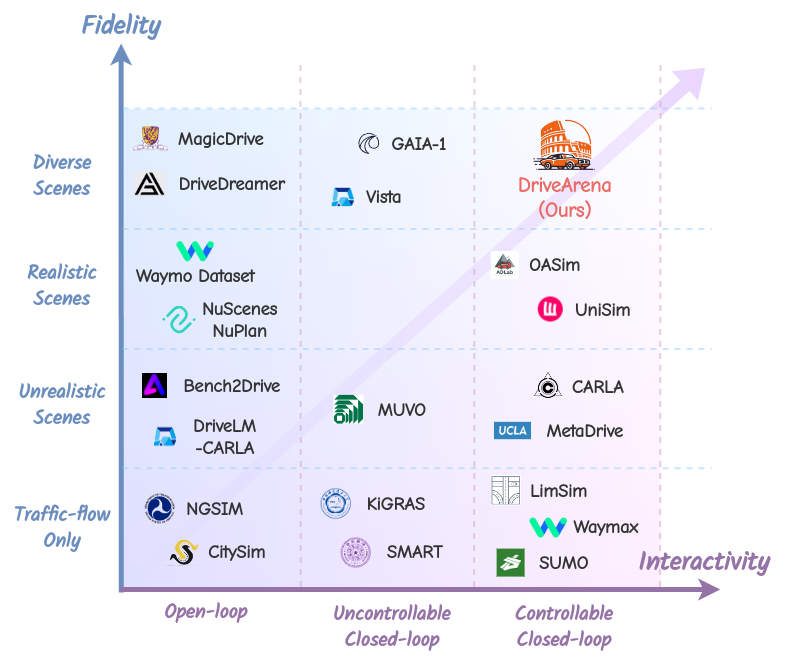}
    \caption{
    Comparison of {\arena} with existing autonomous driving methods and platforms along the dimensions of Interactivity and Fidelity. 
    \textit{Interactivity} indicates the platform's control over vehicles, ranging from open-loop, uncontrollable closed-loop, to controllable closed-loop.
    \textit{Fidelity} reflects the realism of driving scenarios, categorized from bottom to top as: traffic-flow only, unrealistic scenes, realistic scenes, and diverse scenes. 
    {\arena} uniquely occupies the top-right, being the first simulation platform to generate diverse traffic scenarios and surround-view images with closed-loop controllability for all vehicles.
    For detailed descriptions of these methods, please refer to Table~\ref{tab:comparison}.
    }
    \label{fig:wallpaper}
    \vspace{-10pt}
\end{figure}

\section{Introduction}
\label{sec:intro}

Autonomous driving (AD) algorithms have advanced rapidly in recent decades~\cite{ayoub2019manual, chen2023milestones, xing2021toward, ma2023detzero, yang2021semantic, mei2023camera, mei2023centerlps, mei2023ssc, mei2024lidar}, progressing from modular pipelines~\cite{yin2021center, guo2023scenedm, li2023logonet, li2022homogeneous} to end-to-end models~\cite{hu2023planning, ye2023fusionad, jiang2023vad} and knowledge-driven methods~\cite{li2023towards,wen2023dilu,fu2024drive}. Despite demonstrating outstanding performance across various benchmarks, significant challenges persist in evaluating these algorithms on replayed open-loop datasets, obscuring their real-world efficacy.
Public datasets~\cite{caesar2020nuscenes, caesar2021nuplan, sun2020scalability}, while offering realistic driving data with authentic sensor inputs and traffic behavior, are inherently biased towards simple straight-ahead scenarios. In such cases, an agent can achieve seemingly good performance by merely maintaining its current state, complicating the assessment of actual driving capabilities in complex situations. 
Furthermore, the agent's current decision does not affect execution or subsequent decisions in the open-loop evaluation, which prevents it from reflecting cumulative errors in real-world driving scenarios.
Additionally, the static nature of recorded datasets, where other vehicles cannot react to the ego vehicle's behavior, further hinders the evaluation of AD algorithms in dynamic, real-world conditions.

As illustrated in Figure~\ref{fig:wallpaper}, we analyze existing AD methods and platforms, revealing that most of them are inadequate for a high-fidelity closed-loop simulation.
Ideally, as an aspect of embodied intelligence, agents should be evaluated in a closed-loop environment, where other agents react to the actions of the ego vehicle, and the ego vehicle receives changed sensor input accordingly. However, existing simulation environments either cannot simulate sensor inputs~\cite{wenl2023limsim, krajzewicz2012recent, gulino2024waymax} or have a significant domain gap with the real world~\cite{dosovitskiy2017carla, li2022metadrive}, making it difficult to seamlessly integrate algorithms into the real world, thus posing a huge challenge for closed-loop evaluation. 
We believe that the simulator should not only closely reflect the visual and physical aspects of the real world, but also promote the continuous learning and evolution of the model within an exploratory closed-loop system for adapting to diverse complex driving scenarios. To achieve this goal, it is imperative to establish a high-fidelity simulator that complies with physical laws and supports interactive functionalities. 

Therefore, we present {\arena}, a pioneering closed-loop simulator based on conditional generative models for training and testing driving agents. 
Specifically, {\arena} offers a flexible platform that can be integrated with any camera-input driving agent. It adopts a modular design and naturally supports iterative upgrades of each module. {\arena} consists of a {\engine} that manages traffic flow and a {\dreamer} based on auto-regressive generation.
{\engine} can generate realistic interactive traffic flow on any road network worldwide, while {\dreamer} is a high-fidelity conditional generative model with infinite autoregression. The driving agent should make corresponding driving actions based on the images generated by {\dreamer}, and feed them back to {\engine} to update the status of vehicles in the environment. The new scene layout will be returned to {\dreamer} for a new round of simulation. This iterative process realizes the dynamic interaction between the driving agent and the simulation environment. The specific contributions are as follows:

\begin{itemize}
[align=right,itemindent=0em,labelsep=5pt,labelwidth=0em,leftmargin=*,itemsep=0em] 
    \item \textbf{High-fidelity Closed-loop Simulation}: We propose the first high-fidelity closed-loop simulator for autonomous driving,  {\arena}, which can provide realistic surround images and integrate seamlessly with existing vision-based driving agents. It can closely reflect the visual and physical properties of the real world, enabling agents to continuously learn and evolve in a closed-loop manner and adapt to various complex driving scenarios.
    
    \item \textbf{Controllability and Scalability}: 
    Our {\engine} can dynamically control the movement of all vehicles in the scenarios and feed the road and vehicle layouts into {\dreamer}, which utilizes a conditional diffusion framework to generate realistic images in a stable and controllable manner. Additionally, {\arena} supports simulation using road networks from any city worldwide, enabling the creation of diverse driving scenario images with varying styles. 
    
    \item \textbf{Modularized Design}: The Driving Agent, {\engine} and {\dreamer} communicate via network interfaces, enabling a highly flexible and modular framework. This architecture allows each component to be replaced with different methods without requiring specific implementations. Functioning as an \textit{arena} for these \textit{players}, {\arena} facilitates comprehensive testing and improvement of both vision-based autonomous driving algorithms and driving scene generative models.
\end{itemize}

\begin{figure*}[htbp]
    \centering
    \includegraphics[width=0.78\linewidth]{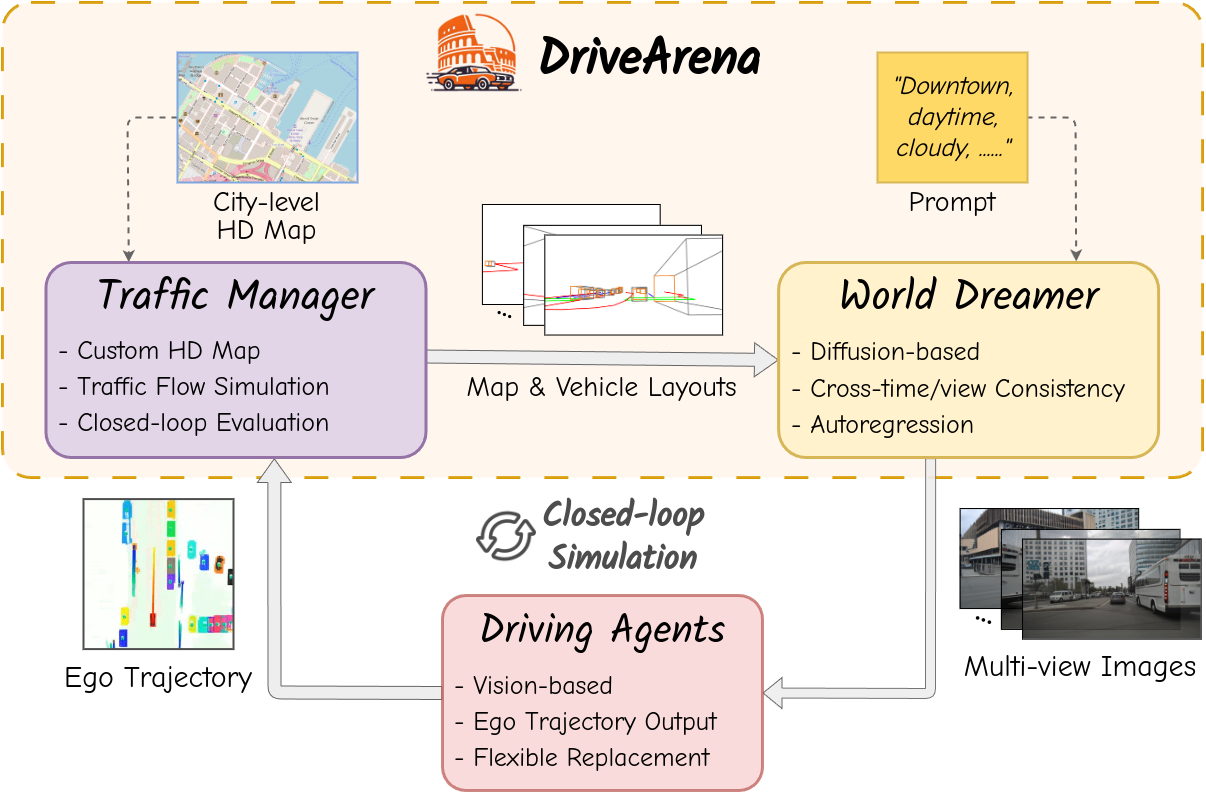}
    \caption{Overview of the {\arena} framework. The system consists of two main components: (1) The {\engine}, which processes Internet-downloaded HD maps to create diverse urban layouts, manages vehicle movements including background traffic, and handles collision detection. (2) The {\dreamer}, an auto-regressive generative model that generates photo-realistic, multi-view camera images corresponding to the simulation state, with controllable parameters following given prompts.
    The framework operates in a closed loop: generated images are fed to the AD agent, which outputs the planned ego trajectory. The trajectory is then fed back into the {\engine} for the next simulation step.}
    \label{fig:pipeline}
\end{figure*}

\section{{\arena} Framework}

As illustrated in Figure~\ref{fig:pipeline}, the framework of our proposed {\arena} comprises two key components: a {\engine} functioning as the backend physical engine and a {\dreamer} serving as the real-world image renderer. Unlike conventional approaches, {\arena} does not rely on pre-built digital assets or reconstructed 3D road models. Instead, the {\engine} adapts to road networks of any city in OpenStreetMap (OSM) format~\cite{haklay2008openstreetmap}, which can be directly downloaded from the Internet. This flexibility enables closed-loop traffic simulations on diverse urban layouts.

The {\engine} receives ego trajectories output by the autonomous driving agent and manages the movement of all background vehicles. Unlike world model approaches \cite{gao2023magicdrive,hu2023gaia} that rely on diffusion models for both image generation and vehicle movement prediction, our {\engine} utilizes explicit traffic flow generation algorithms \cite{wen2023bringing}. This approach enables the generation of a wider range of uncommon and potentially unsafe traffic scenarios, while also facilitating real-time collision detection between vehicles.

{\dreamer} generates realistic camera images that precisely correspond to the {\engine}'s output. It also allows for user-defined prompts to control various elements of the generated images, such as street view style, time of day, and weather conditions, enhancing the diversity of the generated scenes. Specifically, it employs a diffusion-based model that utilizes the current map and vehicle layouts as control conditions to produce surround-view images. These images serve as input for end-to-end driving agents. Given {\arena}'s closed-loop architecture, the diffusion model is required to maintain both cross-view and temporal consistency in the generated images.

The generated multi-view images of the current frame are fed into the end-to-end autonomous driving agents, which can output the ego vehicle's movement. The planned ego trajectory is subsequently sent to {\arena} for the next simulation step. The simulation concludes when the ego vehicle either successfully completes the entire route, crashes, or deviates from the road. Upon completion, {\arena} performs a comprehensive evaluation process to assess the driving agent's capabilities.

It is noteworthy that {\arena} employs a distributed modular design. The {\engine}, {\dreamer}, and AD agent communicate via network using standardized interfaces. Consequently, {\arena} does not mandate specific implementations for the {\dreamer} or the AD agent. Our framework aims to function as an ``\textit{arena}'' for these ``\textit{players}'', facilitating comprehensive testing and improvement of both end-to-end autonomous driving algorithms and realistic driving scene generative models.

\section{Methodology}
\label{sec:method}

Following the {\arena} framework outlined above, we have implemented a preliminary version of {\arena}. In this section, we elaborate on the implementation of each module: {\engine}, {\dreamer}, and AD agent, while describing necessary details that were not previously mentioned. At the end of this section, we present both the open-loop and closed-loop evaluation metrics for AD agents in {\arena}.

\subsection{{\engine}}
Most existing realistic driving simulators \cite{yan2024oasim, yang2023unisim, wu2023mars} rely on limited layouts from public datasets, lacking diversity for dynamic environments. To address these challenges, we utilize LimSim \cite{wenl2023limsim,fu2024limsim++} as the underlying {\engine} to simulate dynamic traffic scenarios and generate road and vehicle layouts for subsequent environment generation. LimSim also provides a user-friendly front-end GUI, which directly displays the BEV map and results from {\dreamer} and the driving agent.

Our {\engine} enables interactive simulations of multiple vehicles in traffic flow, including comprehensive vehicle planning and control. We adopt a hierarchical multi-vehicle decision-making and planning framework, which jointly makes decisions for all vehicles within the flow and reacts promptly to the dynamic environment through a high-frequency planning module~\cite{wen2023bringing}. The framework also incorporates a cooperation factor and trajectory weight set, introducing diversity to autonomous vehicles in traffic at both social and individual levels.

Furthermore, our dynamic simulator supports various custom HD maps of any city from OpenStreetMap, facilitating the construction of diverse road graphs for convenient simulation. 
The {\engine} controls the movement of all background vehicles. For the ego vehicle, we provide two distinct simulation modes: open-loop and closed-loop. In closed-loop mode, the driving agent performs planning for the ego vehicle, and {\engine} uses the agent-outputted trajectory to control the ego vehicle accordingly. In open-loop mode, the trajectory generated by the driving agent is not actually used to control the ego vehicle; instead, {\engine} maintains control in a closed-loop manner. The details of these two modes are further elaborated in Section~\ref{sec: simu modes}.

\subsection{{\dreamer}}

\begin{figure*}[htbp]
    \centering
    \includegraphics[width=0.9\linewidth]{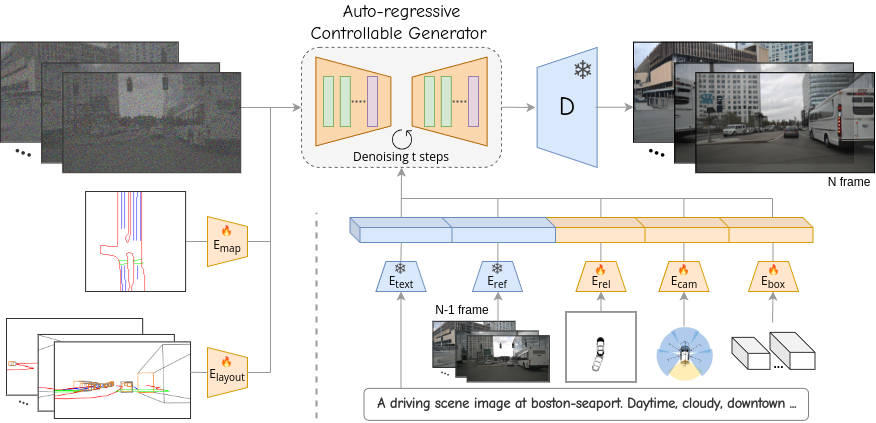}
    \caption{The figure illustrates the denoising process employed by {\dreamer}. Beginning with randomly sampled noise, the autoregressive model utilizes various conditions—such as multi-view layout, BEV map, text prompt, reference image, relative pose, camera parameters, and 3D bounding boxes—to enhance the denoising procedure. The encoders depicted in the figure are distinct, with the color indicating whether each one utilizes a pre-trained network and is frozen. Additionally, we incorporate ControlNet to introduce conditional control into the diffusion model.}
    \label{fig:generator}
\end{figure*}
Unlike recent autonomous driving generation methods \cite{yan2024oasim, yang2023unisim, wu2023mars} that use Neural Radiance Fields (NeRF) and 3D Gaussian Splatting for environment reconstruction from logged video, we design a diffusion-based {\dreamer}. It utilizes control conditions of the map and vehicle layouts from the {\engine} to generate geometrically and contextually accurate driving scenarios. Our framework shares several advantages: (1) Better controllability. The generated scenes can be controlled by scene layouts from {\engine}, textual prompts, and reference images to capture different weather conditions, lighting, and scene styles. (2) Better scalability. Our framework can adapt to various road structures without the need to model the scene in advance. In theory, we support the generation of driving scenes for any city in the world by leveraging layouts from OpenStreetMap.

We illustrate our diffusion-based {\dreamer} in Figure \ref{fig:generator}. Built upon the stable diffusion pipeline \cite{blattmann2023stable}, {\dreamer} utilizes an effective condition encoding module that accepts a variety of conditional inputs including map and vehicle layouts, text descriptions, camera parameters, ego poses, and reference images to generate realistic surround-view images. Considering the importance of ensuring synthesis scene consistency across different views and time spans for driving agents, we integrate a cross-view attention module, inspired by~\cite{gao2023magicdrive}, to maintain coherence across different views. Additionally, we adopt an image auto-regressive generation paradigm to enforce temporal consistency. This approach enables {\dreamer} to not only maximally maintain the temporal consistency of the generated videos, but also generate videos of arbitrary length in an infinite stream, which provides great support for autonomous driving simulation.

\noindent\textbf{Condition encoding.} Previous work \cite{gao2023magicdrive} applied BEV layout as conditional input to control the output of the diffusion model, which increased the difficulty of the network in learning to generate geometrically and contextually accurate driving scenes. In this work, we present a new condition encoding module to introduce more guidance information, which helps the diffusion module generate high-fidelity surround images. Specifically, in addition to encoding camera poses for each view, text descriptions, 3D object bounding boxes, and BEV map layouts using a condition encoder similar to \cite{gao2023magicdrive}, we also explicitly project the map and object layouts onto each camera view to generate layout canvases for more accurate lane and vehicle generation guidance. 
Specifically, the text embedding $e_{text}$ is obtained by encoding the text descriptions with the CLIP text encoder~\cite{radford2021learning}. The parameters $\mathbf{P} = \{ \mathbf{K} \in \mathbb{R}^{3 \times 3}, \mathbf{R} \in \mathbb{R}^{3 \times 3}, \mathbf{T} \in \mathbb{R}^{3 \times 1} \}$ of each camera and the 8 vertices of the 3D bounding boxes are encoded to $e_{cam}$ and $e_{box}$ by Fourier embedding~\cite{mildenhall2021nerf}, where $\mathbf{K}$, $\mathbf{R}$, $\mathbf{T}$ represent camera intrinsic, rotations and translations respectively. The 2D BEV map grid uses the same encoding method as in ~\cite{gao2023magicdrive} to get embedding $e_{map}$. Then, each category of the HD maps and the 3D boxes is projected onto the image plane to obtain the map canvas and box canvas, respectively. These canvases are concatenated to create the layout canvas. The final feature $e_{layout}$ can be obtained by encoding the layout canvas by the conditional encoding network~\cite{zhang2023adding}.

Moreover, we introduce a reference condition to provide appearance and temporal consistency guidance. During training, we randomly extract a frame from the past $\mathbf{L}$ frames as a reference frame and use the pre-trained CLIP model \cite{radford2021learning} to extract reference features $e_{ref}$ from the multi-view images. The encoded reference features imply semantic context and are integrated into the conditional encoder through the cross-attention module. In order to make the diffusion model aware of the motion changes of the ego-vehicle, we also encode the ego-pose relative to the reference frame into the conditional encoder to capture the motion change trend of the background. The relative pose embedding $e_{rel}$ is encoded by Fourier embedding.
By incorporating the above control conditions, we can effectively control the generation of surround images.

\noindent\textbf{Auto-regressive generation.} To facilitate online inference and streaming video generation while maintaining temporal coherence, we have developed an auto-regressive generation pipeline. Specifically, during the inference phase, the previously generated images and the corresponding relative ego pose are used as reference conditions. This approach guides the diffusion model to generate current surround images with enhanced consistency, ensuring a smoother transition and coherence with the previously generated frames.

What we designed in this paper is just a simple implementation of the {\dreamer}. We also verify that extending the auto-regressive generation to a multi-frame version (using multiple past frames as references and outputting multi-frame images) and adding additional temporal modules can improve temporal consistency.

\subsection{Driving Agent}
Recent works \cite{li2024ego, zhai2023rethinking} have demonstrated the challenges in justifying the planning behavior of driving agents through open-loop evaluation on public datasets \cite{caesar2020nuscenes}, primarily due to the simplistic nature of driving scenarios presented. While some studies \cite{wang2023drivemlm} have conducted closed-loop evaluations using simulators like CARLA \cite{dosovitskiy2017carla}, discrepancies such as appearance and scene diversity persist between these simulations and the dynamic real world. To bridge this gap, our {\arena} provides a realistic simulation platform with the corresponding interfaces for camera-based driving agents \cite{jiang2023vad, hu2023planning,hu2022st} to perform more comprehensive evaluations, including both open-loop and closed-loop testing. Moreover, by changing the input conditions, such as the road and vehicle layouts, {\arena} could generate corner cases and facilitate these driving agents' evaluation on out-of-distribution scenarios. Without loss of generality, we select a representative end-to-end driving agent, namely UniAD \cite{hu2023planning}, to conduct both open-loop and closed-loop testing in our {\arena}. UniAD utilizes surround images to predict motion trajectories for the ego vehicle and other agent vehicles, which can be seamlessly integrated with the API of our dynamic simulator for evaluation. Furthermore, the perceptual outputs, such as 3D detection and map segmentation, contribute to enhancing the validation of realism in our environment generation.

\begin{figure*}[htbp]
    \centering
    \includegraphics[width=0.95\linewidth]{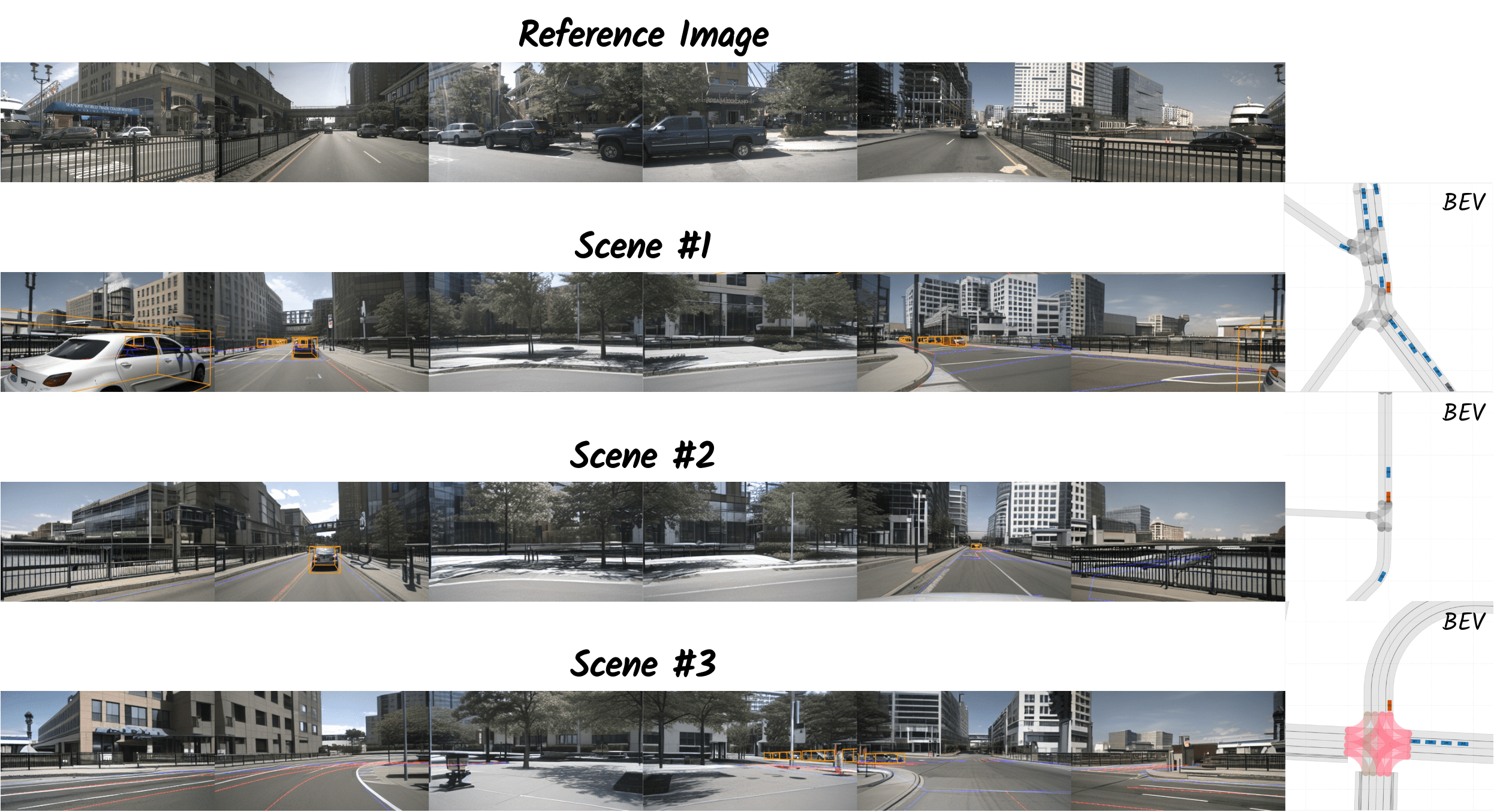}
    \caption{Demonstration of reference image influence on generated scenes. Three scenes are presented, all derived from a single nuScenes reference frame. Despite notable variations in road networks, {\dreamer} successfully integrates street styles and weather conditions from the reference image while adhering to specified control conditions for vehicles and road layouts. Of particular interest is the aerial corridor visible in the reference image, which is accurately reproduced in scenes \#1 and \#2. However, in scene \#3, due to the curved road configuration, the corridor is not generated, illustrating {\dreamer}'s adaptability to different road geometries.}
    \label{fig:same_ref}
    \vspace{-10pt}
\end{figure*}

\subsection{Ego Control Modes and Evaluation Metrics}
\label{sec: simu modes}

{\arena} inherently supports ``closed-loop'' simulation mode of driving agents. That is, the system adopts the trajectory output by the agent at each timestep, updates the ego vehicle's state based on this trajectory, and simulates the actions of background vehicles. Subsequently, it generates multi-view images for the next timestep, thus maintaining a continuous feedback closed-loop.
Additionally, recognizing that some AD agents may be unable to perform long-term closed-loop simulation during the development process, {\arena} also supports the ``open-loop'' simulation mode. In this mode, the {\engine} will take over the control of the ego vehicle, while the trajectory output by the AD agent is recorded for subsequent evaluation.

In both open-loop and closed-loop modes, it is crucial to comprehensively evaluate AD agent performance from a results-oriented perspective. Drawing inspiration from NAVSIM~\cite{Dauner2024navsim} and the CARLA Autonomous Driving Leaderboard~\cite{carla2023leaderboard}, {\arena} adopts two evaluation metrics: PDM Score (PDMS) and Arena Driving Score (ADS).

PDMS, initially proposed by NAVSIM~\cite{Dauner2024navsim}, evaluates the trajectory output at each timestep. We adhere to the original definition of PDMS, which aggregates the following sub-scores:
\small
\begin{equation}
\begin{aligned}
\mathrm{PDMS}_t=&\underbrace{\left(\prod_{m \in\{\mathrm{NC}, \mathrm{DAC}\}} \text { score }_m\right)}_{\text {penalties}} \times \\ &\underbrace{\left(\frac{\sum_{w \in\{\mathrm{EP}, \mathrm{TTC}, \mathrm{C}\}} \mathrm{weight}_w  \times \text { score }_w}{\sum_{w \in\{\mathrm{EP}, \mathrm{TTC}, \mathrm{C}\}} \mathrm{weight}_w}\right)}_{\text {weighted average }}.
\end{aligned}
\end{equation}
\normalsize
where the penalties include the drive with no collisions (NC) with road users and drivable area compliance (DAC), as well as the weighted average including ego progress (EP), time-to-collision (TTC), and comfort (C). 
We implement minor modifications tailored to {\arena}: in $\text{score }_\mathrm{NC}$, we do not differentiate ``at-fault" collisions, and for $\text{score }_\mathrm{EP}$, we utilize the {\engine}'s Ego path planner as the reference trajectory instead of the Predictive Driver Model.
At the end of the simulation, the final PDM Score is averaged across all simulation frames.
\begin{equation}
\mathrm{PDMS} = \frac{\Sigma_{t=0}^{T}{\mathrm{PDMS}_t}}{T} \in [0,1]
\end{equation}

For open-loop simulations, PDMS serves directly as the evaluation metric for AD agents. However, for driving agents operating under the ``closed-loop" simulation mode, we employ a more comprehensive metric called Arena Driving Score (ADS), which combines the trajectory PDMS with route completion:
\begin{equation}
\mathrm{ADS} = \mathrm{R}_{c} \times \mathrm{PDMS}
\end{equation}
where $\mathrm{R}_{c} \in [0,1]$ represents route completion, defined as the percentage of the route distance completed by an agent. Given that ``closed-loop" simulations terminate upon agent collision with other road users or deviation from the road, ADS provides a suitable metric for differentiating agents' driving safety and consistency.

\section{Experiments}
\label{sec:exp} 

\subsection{{\dreamer} Setups}
\noindent \textbf{Dataset.} 
For {\dreamer}, we use the nuScenes~\cite{caesar2020nuscenes} dataset for training. Following the official configuration, we employ 700 scenes for training and 150 for validation. We focus on four road categories (lane boundary, lane divider, pedestrian crossing, and drivable area) and ten object categories. The nuScenes dataset contains data collected from four different cities, covering various light and weather conditions, including daytime, night, sunny, cloudy, and rainy scenarios, enabling {\arena} to conditionally imitate diverse appearances. We additionally annotated each scene using GPT-4V, providing detailed scene descriptions that include elements like time, weather, street style, road structure, and appearance. These descriptions serve as text prompt conditions.

\begin{figure*}[htbp]
    \centering
    \includegraphics[width=0.99\linewidth]{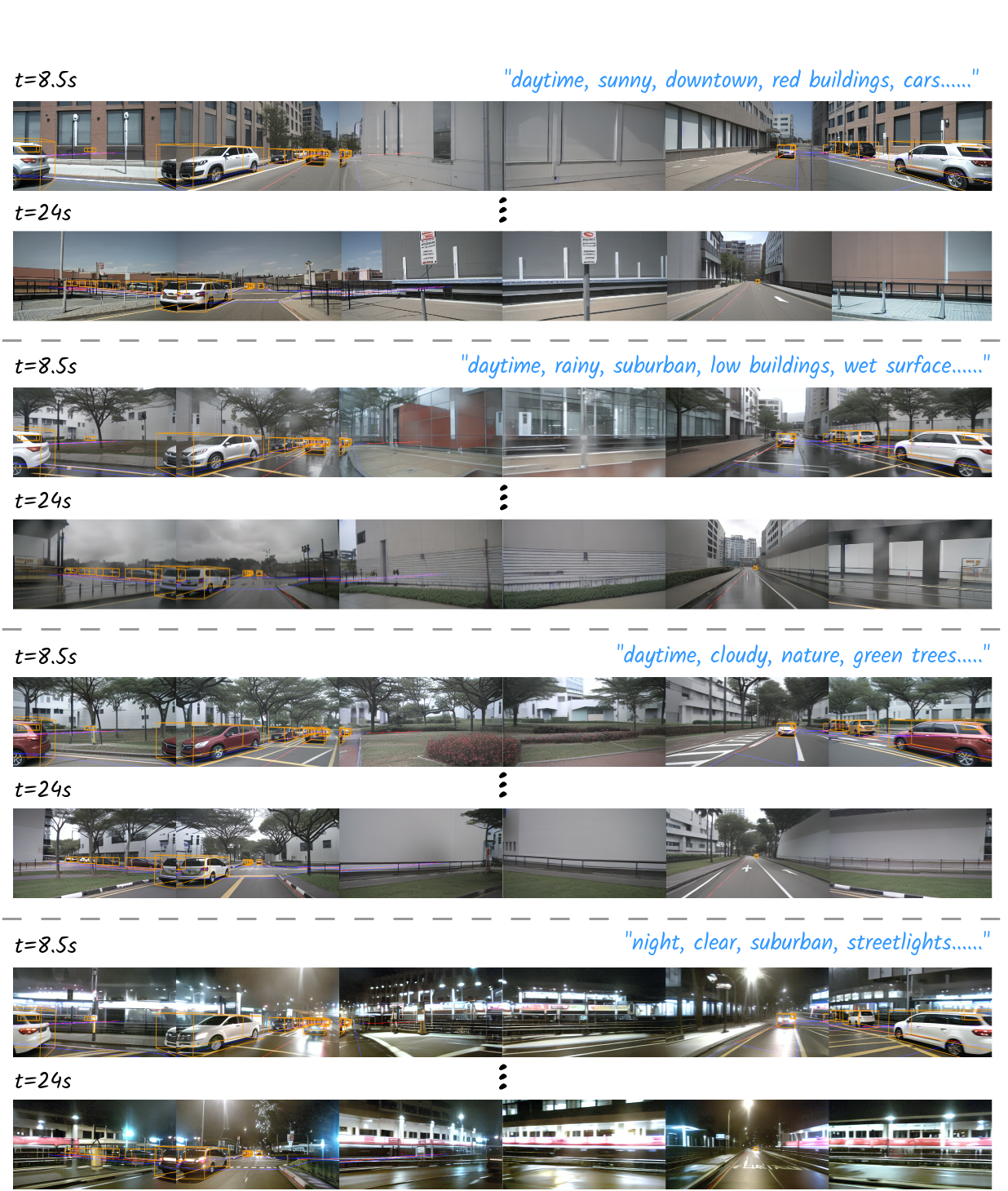}
    \caption{Demonstration of diverse prompts and reference images' influence on identical scenes. The figure presents four distinct image sequences generated by {\arena} for a same 30-second simulation sequence, each utilizing different prompts and reference images. All sequences strictly adhere to the provided control conditions for road structures and vehicles, maintaining cross-view consistency. Notably, the four sequences exhibit significant variations in weather and lighting conditions, while consistently preserving their respective styles throughout the entire 30-second duration.  \href{https://pjlab-adg.github.io/DriveArena/}{Click here for video demonstration.}}
    \label{fig:weather}
\end{figure*}

\begin{table*}[htbp]
\setlength{\tabcolsep}{2.5pt}
\small
\centering
\begin{tabular}{l|cc|cccc|cccc|cccc} 
\toprule
\multirow{2}{*}{Source of test data} & \multicolumn{2}{c|}{3DOD} & \multicolumn{4}{c|}{BEV Segmentation mIoU (\%)} & \multicolumn{4}{c|}{L2 (m) $\downarrow$}& \multicolumn{4}{c}{Col. Rate (\%) $\downarrow$} \\ 
\cmidrule{2-15}
                                     & mAP $\uparrow$  & NDS $\uparrow$ & Lanes $\uparrow$ & Drivable $\uparrow$ & Divider $\uparrow$ & Crossing $\uparrow$ & 1.0s & 2.0s & 3.0s & Avg.  & 1.0s   & 2.0s   & 3.0s   & Avg.     \\ 
\cmidrule{1-15}
ori nuScenes & \textbf{37.98} & \textbf{49.85} & \textbf{31.31} & \textbf{69.14}    & \textbf{25.93} & \textbf{14.36} & \textbf{0.51} & \textbf{0.98} & \textbf{1.65} & \textbf{1.05}  & \underline{0.10} & \textbf{0.15} & \underline{0.61} & \underline{0.29}  \\
MagicDrive   & 12.92 & 28.36 & 21.95 & 51.46 & 17.10 & 5.25 & 0.57 & 1.14 & 1.95 & 1.22  &  \underline{0.10} & 0.25 & 0.70 & 0.35  \\
{\textbf{\arena}} & \underline{16.06} & \underline{30.03} & \underline{26.14} & \underline{59.37} & \underline{20.79} & \underline{8.92} & \underline{0.56} & \underline{1.10} & \underline{1.89} & \underline{1.18}  & \textbf{0.02} & \underline{0.18} & \textbf{0.53} & \textbf{0.24}  \\
\bottomrule
\end{tabular}
\caption{Comparison of generation fidelity. The data synthesis conditions are from the nuScenes validation set. All results are computed by using the official implementation and checkpoints of UniAD. \textbf{Bold} represents the best results, \underline{underline} represents the second best results.}
\label{tab:metrics}
\end{table*}

\noindent \textbf{Model Setup.} 
The model is initialized with the pre-trained Stable Diffusion v1.5~\cite{rombach2022high}, with only the newly added parameters being trained. For various conditions, except for the encoding of reference images and text prompts, the encoders for other conditions are randomly initialized and trained from scratch. These conditions are then integrated into the UNet using a randomly initialized ControlNet~\cite{zhang2023adding} to control the denoising process.

\noindent \textbf{Training and Inference.} 
To utilize the reference images and achieve temporal correlation, we employ ASAP~\cite{wang2023we} to generate 12Hz interpolated annotations and crop them into image clips of length $\mathbf{L}=7$. During training, we use the last frame of each clip as the current frame, select any frame from the clip as the reference frame, and calculate the relative pose between them to model the motion trend of the background. Accordingly, the surround images corresponding to the reference frame are input to the network as reference images. During inference, the generated result of the previous frame is used as the current reference images, enabling unlimited length generation. 
The experiment is conducted on 8 NVIDIA A100 (80GB) GPUs with a batch size of 4×8 and 200K iterations of training. The AdamW optimizer is used with a learning rate of 1e-4.
The network follows the same image resolution (224×400) as MagicDrive, and when input to the driving agent, it will be upsampled to the original image size of nuScenes (900×1600) through a super-resolution algorithm~\cite{wang2024camixersr}.

\subsection{{\engine} Setups}

\noindent \textbf{Operating Frequencies.}
In our experiments, the {\engine} operates at a frequency of 10Hz, while the control frequency is set to 2Hz. This configuration results in the {\engine} sending the current layout to {\dreamer} every 0.5 simulation seconds, requesting surround images. These images are then forwarded to the driving agent, which predicts and plans the subsequent trajectory for the ego vehicle. The {\engine}, {\dreamer}, and driving agent communicate via HTTP protocol, enabling deployment across different servers.

\noindent \textbf{Simulation Modes.}
As detailed in Section~\ref{sec: simu modes}, we implement two simulation modes. In the open-loop mode, all vehicles, including the ego vehicle, are controlled by {\engine} itself. The driving agent can predict the ego vehicle's trajectory, but its trajectory is not actually executed. In the closed-loop mode, the ego vehicle is controlled by the driving agent, and the simulation terminates if it crashes with other vehicles or leaves the road.

\noindent \textbf{Supported Maps.}
Currently, {\arena} supports four different maps, which are: \texttt{singapore-onenorth}, \texttt{boston-seaport}, \texttt{boston-thomaspark}, and \texttt{carla-town05}. The first two maps closely resemble the corresponding areas in the nuScenes dataset, while the last one replicates the road network of the Town05 map in the CARLA simulator. Notably, {\engine} can download road network data for any area directly from OpenStreetMap and perform simulations, enabling {\arena} to simulate the road network of almost any city worldwide. OpenstreetMap also accepts customized operations, and users can draw the desired road network structure for simulation testing.

\subsection{{\dreamer} Fidelity Validation}
To assess the sim-to-real gap between our generated images and the original nuScenes images, we employ UniAD~\cite{hu2023planning} as an evaluator. We generate videos for 150 scenes based on the original layout provided by the nuScenes validation set with 2Hz. For comparative analysis, we set MagicDrive as the baseline method and use its official codes and checkpoints for inference. Subsequently, UniAD is performed on these images to compute various metrics, including 3d object detection, BEV map segmentation, and planning. The results are summarized in Table~\ref{tab:metrics}. It shows that all our indicators are higher than the baseline method, and a few indicators even surpass the performance on the original nuScenes. Furthermore, it demonstrates our model's superior capability to accurately respond to control signals and strictly adhere to input conditions. These findings establish a solid foundation for using our generator as a reliable simulator.

\begin{figure*}[htbp]
    \centering
    \includegraphics[width=1.0\linewidth]{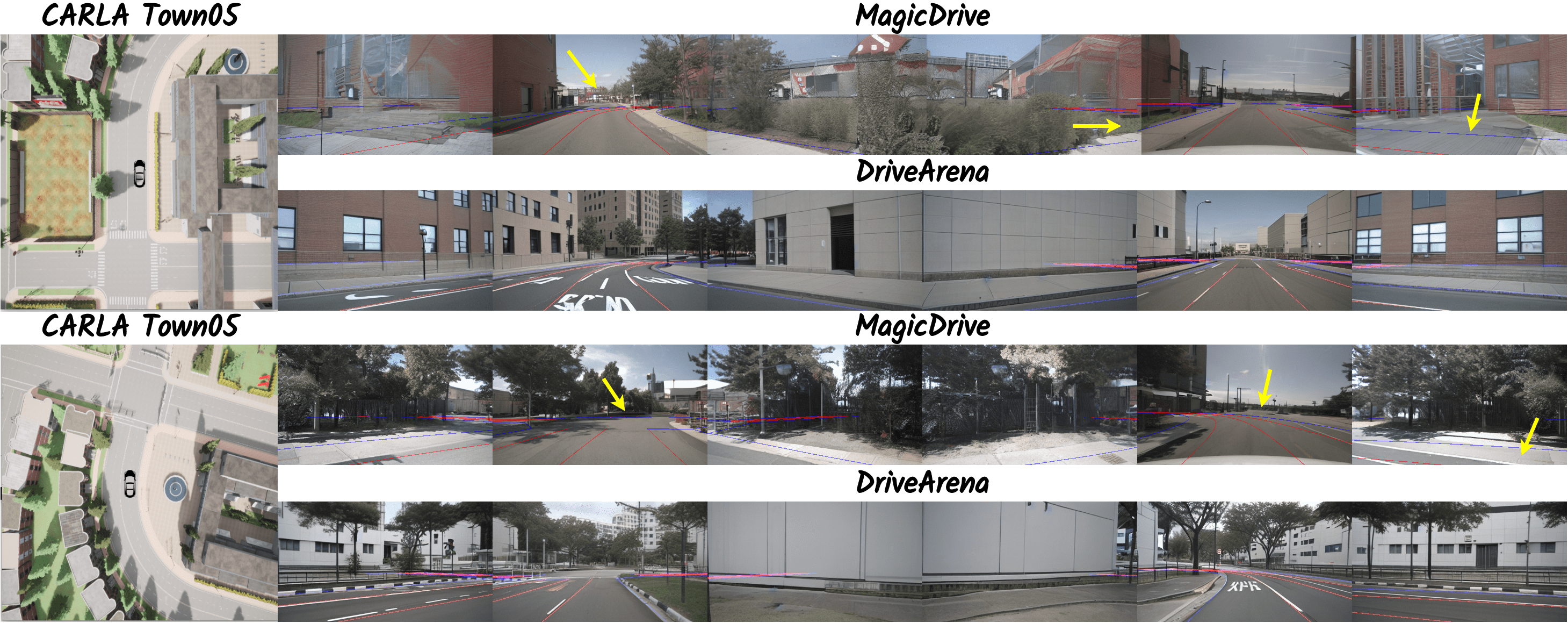}
    \caption{Comparison between MagicDrive and {\arena}. Both are used to generate realistic images on the same Carla Town05 Map, with corresponding ground truth lane lines projected onto the images for demonstration. For such large curvatures and wide roads in CARLA, which are atypical scenarios in nuScenes dataset, MagicDrive struggles to generate images that accurately fit the network. It incorrectly generates pavements and fails to match the road curvature (indicated by yellow arrows). In contrast, {\arena} successfully generates images that accurately represent the road structure.}
    \label{fig:md_cmp3}
\end{figure*}

\begin{figure*}[htbp]
    \centering
    \includegraphics[width=\linewidth]{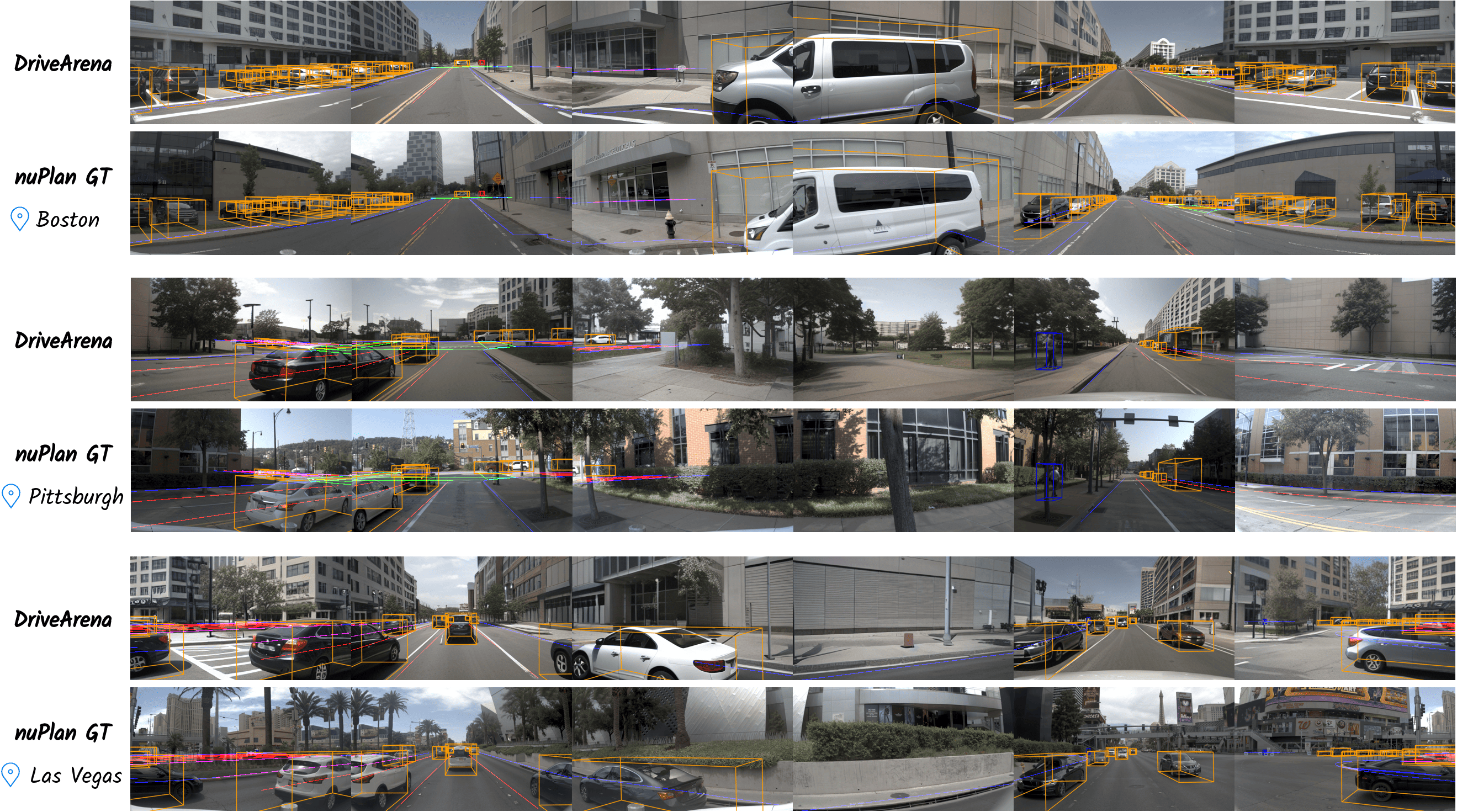}
    \caption{Zero-shot inference on nuPlan datasets. {\dreamer}, trained exclusively on the nuScenes dataset, demonstrates remarkable adaptability when applied to the nuPlan dataset. The latter comprises data from new cities (Pittsburgh, Las Vegas) not present in nuScenes, with different camera configurations and parameters. We selected three nuPlan scenes and directly utilized nuPlan's camera parameters to project object boxes and lane lines onto the corresponding images as control conditions. The results show that {\dreamer} produces coherent images when deployed in unfamiliar cities and even with previously unseen camera configurations and layouts.}
    \label{fig:nuplan_exp}
\end{figure*}

\subsection{Visualization}
\noindent\textbf{Controllability.}
In this section, we will comprehensively demonstrate the controllability of the model from various dimensions, including the control of lighting and weather, the fit of object boxes and maps, change of street style, and consistency over long periods of time.

We demonstrate the impact of the reference image on the generated image, as shown in Figure~\ref{fig:same_ref}. We randomly select one frame of images from the nuScenes dataset as reference images and choose three scenes from OpenStreetMap and Carla. We perform inference on them with {\dreamer} respectively. It can be seen that the source and style of the road network are very different from the scope of the original nuScenes dataset. The pictures show that the generated vehicles and road networks conform closely to control conditions, demonstrating strong control capabilities. The style and weather of the generated pictures can also be consistent with the reference images. In other words, besides maintaining image generation continuity through reference images, we can also regulate image style accordingly.

Figure~\ref{fig:weather} presents images generated using different text prompts and reference images on the same road network. Each set of images portrays the surrounding scenery at intervals of 8.5 seconds and 24 seconds respectively, with the layout projected on the image. The images clearly 
illustrate that the road structure and vehicles strictly adhere to the given control conditions while maintaining excellent consistency in the surround view. In addition, the four sets of images exhibit significant differences in weather and lighting and can maintain their own styles during the continuous iteration process.

\noindent\textbf{Scalability.}
The {\engine} can accept any map downloaded from OpenStreetMap and seamlessly connect to the Carla road network. Combined with Dreamer's excellent following capability, the entire framework demonstrates robust scalability. The specific results are shown in Figure~\ref{fig:md_cmp3}. We used both MagicDrive and our {\dreamer} to generate realistic images on the same Carla road network, with the corresponding lane lines projected onto the images. The road style in Carla differs significantly from that of nuScenes. It is rare to encounter roads with such large curvature and such wide roads in nuScenes. Consequently, the performance of MagicDrive, which is based on the nuScenes BEV map, is slightly inferior in these conditions. As indicated by the yellow arrow, MagicDrive struggles to generate curved roads and fit wide roads accurately. {\arena}, however, can produce reasonable pictures that follow the road structure.

We also demonstrate additional cases using data from the nuPlan dataset to validate the scalability. The nuPlan data originates from cities different from nuScenes and features varying camera numbers and parameters. We select 6 cameras with a similar layout to the nuScenes dataset, and nuPlan's camera parameters are employed to project object boxes and lane lines onto corresponding images as control conditions. As shown in Figure~\ref{fig:nuplan_exp}, {\dreamer} fully trained on nuScenes adeptly adheres to these conditions, generating coherent images when deployed in new cities and even with novel camera configurations.

\begin{figure*}[htbp]
    \centering
    \includegraphics[width=0.9\linewidth]{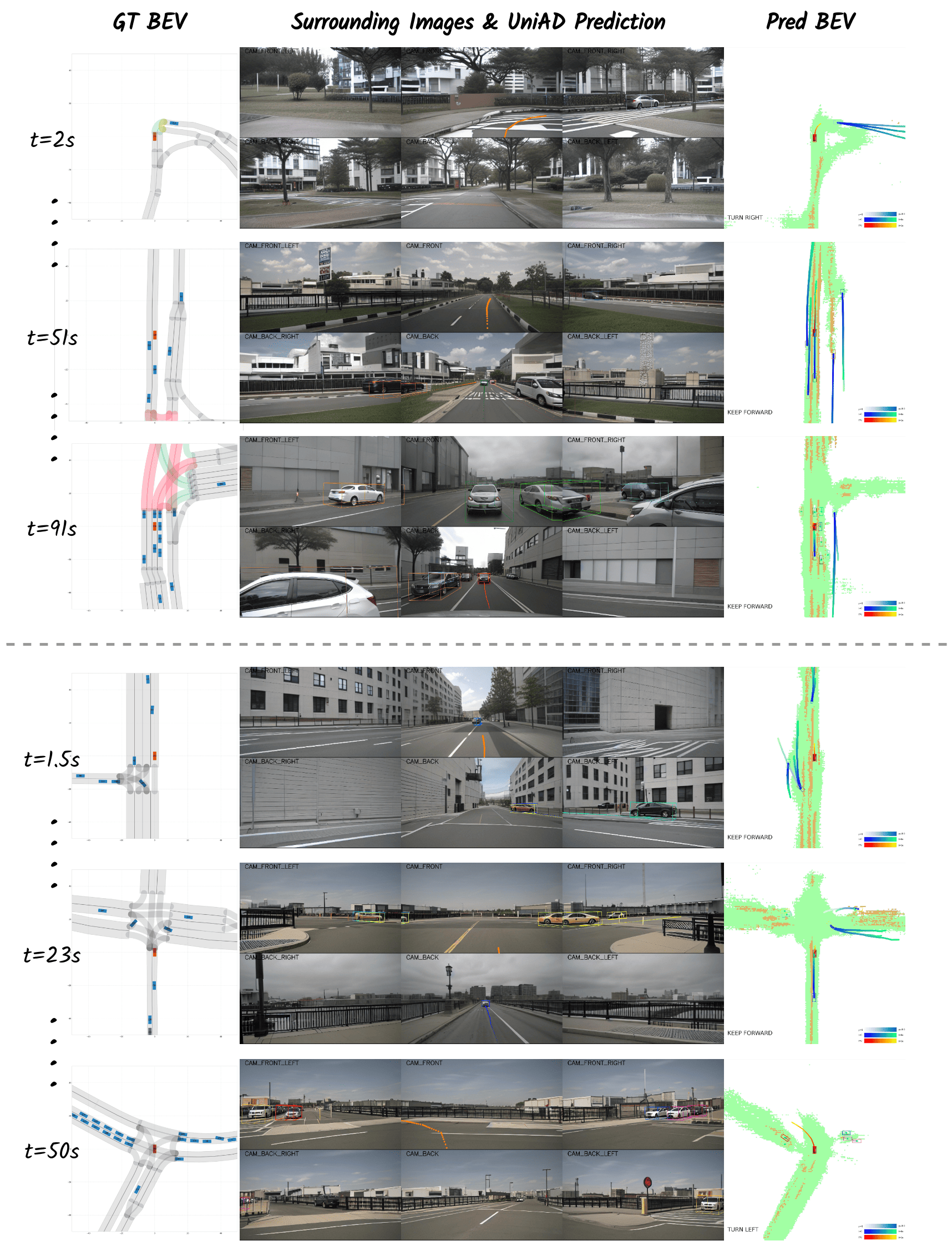}
    \caption{Case studies of UniAD's open-loop performance in {\arena}. The figure presents two long-term open-loop simulation sequences: the upper sequence depicts a Singapore road network and style (left-hand drive), while the lower sequence shows a Boston road network and style (right-hand drive). Each subfigure displays, from left to right: {\engine}'s ground truth BEV; {\dreamer}-generated image with corresponding UniAD detection bounding boxes and predicted trajectories; and UniAD-predicted BEV image. \href{https://pjlab-adg.github.io/DriveArena/}{Click here for video demonstration.}}
    \label{fig:uniad openloop}
\end{figure*}

\begin{figure}[tbp]
    \centering
    \includegraphics[width=1.0\linewidth]{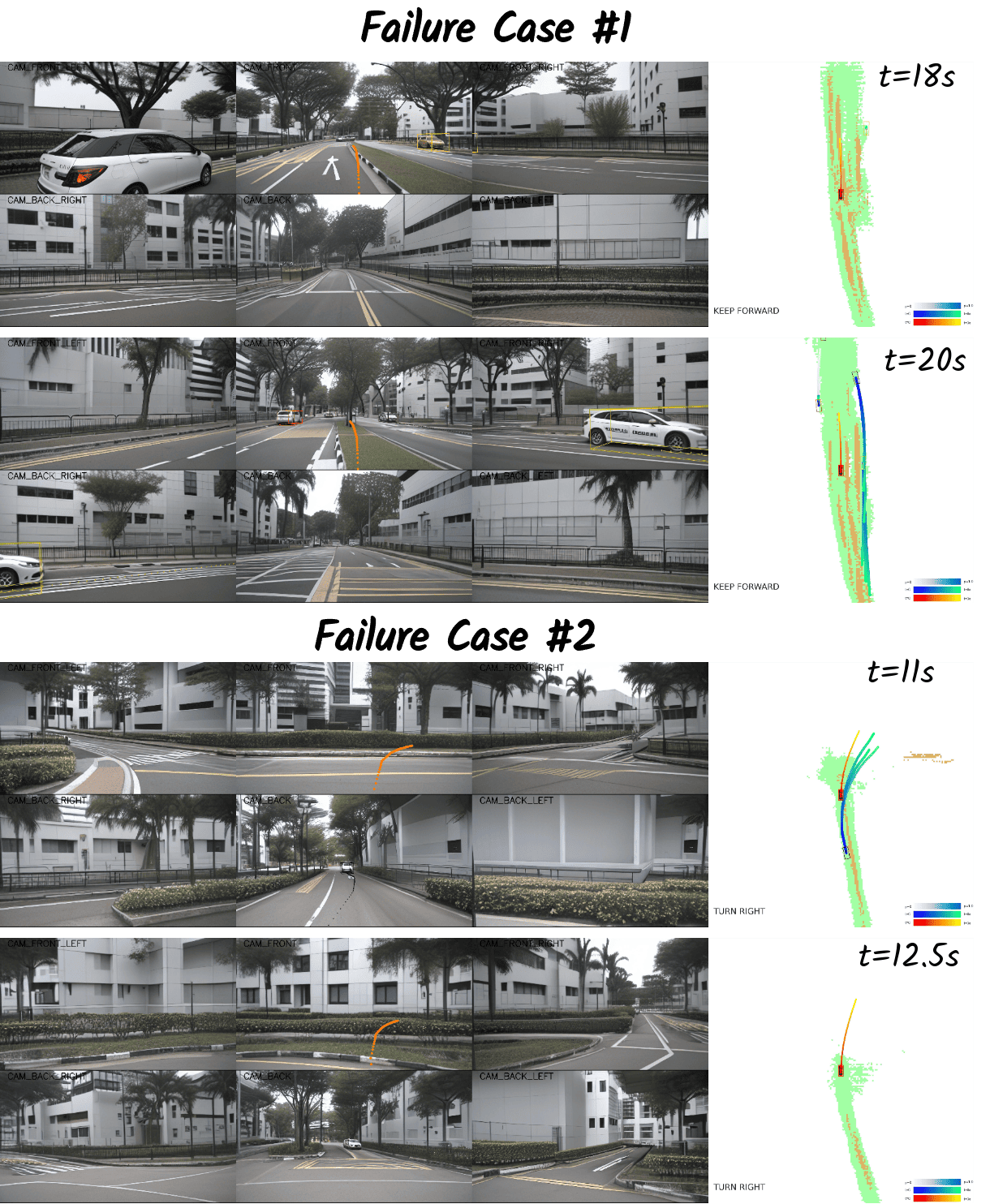}
    \caption{Failure cases of UniAD in {\arena}'s closed-loop mode. While UniAD generally predicts road structures accurately: (top) UniAD encroaching onto the central median; (bottom) UniAD failing to complete a right turn successfully. }
    \label{fig:uniad fail1}
\end{figure}

\begin{table*}[tbp]
\centering
\resizebox{0.85\textwidth}{!}{%
\begin{tabular}{l|cccccc}
\toprule
UniAD perform in & NC $\uparrow$ & DAC $\uparrow$ & EP $\uparrow$ & TTC $\uparrow$ & C $\uparrow$ & PDMS $\uparrow$\\ \midrule
nuScenes (original) & 0.993±0.03 & 0.995±0.01 & 0.914±0.05 & 0.947±0.14 & 0.848±0.21 & 0.910±0.09 \\
nuScenes (generated) & 0.993±0.02 & 0.991±0.02 & 0.909±0.05 & 0.951±0.14 & 0.821±0.21 & 0.902±0.09 \\
{\arena} (open-loop) &  0.792±0.11 & 0.942±0.04 & 0.738±0.11 & 0.771±0.12  & 0.749±0.16 & 0.636±0.08 \\
\midrule
\rowcolor{black!10}Human (nuScenes GT) & 1.000±0.00 & 1.000±0.00 & 1.000±0.00 & 0.979±0.12 & 0.752±0.17 & 0.950±0.06 \\ 
\bottomrule
\end{tabular}%
}
\caption{UniAD performance in {\arena}'s open-loop mode. Evaluation across three scenarios: \textit{1)} original nuScenes images sequnces; \textit{2)} {\dreamer}-generated images with nuScenes ground truth trajectories; and \textit{3)} {\arena}'s open-loop mode simulation sequences. Metrics include: no collisions (NC), drivable area compliance (DAC), ego progress (EP), time-to-collision (TTC), comfort (C), and PDM Score (PDMS).}
\label{tab:open-loop perform}
\vspace{-5pt}
\end{table*}

\begin{table}[tbp]
\centering
\resizebox{0.9\linewidth}{!}{%
\begin{tabular}{c|ccc}
\toprule
 Route & PDMS $\uparrow$& RC $\uparrow$& ADS $\uparrow$\\ \midrule
\texttt{sing\_route\_1} & 0.7615 & 0.1684 & 0.1282 \\
\texttt{sing\_route\_2} & 0.7215 & 0.169 & 0.0875 \\
\texttt{boston\_route\_1} & 0.4952 & 0.091  & 0.0450 \\
\texttt{boston\_route\_2} & 0.6888 & 0.121 & 0.0835 \\
\midrule
\rowcolor{black!10}Avg. & 0.667±0.12 & 0.137±0.04 & 0.086±0.03 \\
\bottomrule
\end{tabular}%
}
\caption{UniAD performance evaluation in {\arena}'s closed-loop mode across four distinct routes. Performance metrics include: PDM Score (PDMS), Route Completion (RC), and Arena Driving Score (ADS).}
\label{tab:uniad closed-loop perform}
\vspace{-5pt}
\end{table}

\begin{table*}[t]
\resizebox{\textwidth}{!}{
\setlength{\tabcolsep}{3pt}
\begin{tabular}{clcccccccc}
\toprule
\multirow{2}{*}{\textbf{Type}} & \multicolumn{1}{c}{\multirow{2}{*}{\textbf{Name}}} & \multicolumn{2}{c}{\textbf{Interactivity}} & \multicolumn{2}{c}{\textbf{Fidelity}} & \multicolumn{4}{c}{\textbf{Diversity}} \\
\cmidrule(lr){3-4} \cmidrule(lr){5-6} \cmidrule(lr){7-10}
 &  & \begin{tabular}[c]{@{}c@{}}Uncontrollable \\ Closed-loop\end{tabular} & \multicolumn{1}{c}{\begin{tabular}[c]{@{}c@{}}Controllable \\ Closed-loop\end{tabular}} & \begin{tabular}[c]{@{}c@{}}Realistic \\ Images\end{tabular} & \multicolumn{1}{c}{\begin{tabular}[c]{@{}c@{}}Real-world \\ Roadgraph\end{tabular}} & \begin{tabular}[c]{@{}c@{}}Different \\ daylight/weather\end{tabular} & \begin{tabular}[c]{@{}c@{}}Multi-view \\ Images\end{tabular} & \begin{tabular}[c]{@{}c@{}}Unlimited \\ Video\end{tabular} & \begin{tabular}[c]{@{}c@{}}Unspecified \\ map\end{tabular} \\ \midrule

\multicolumn{1}{c|}{\multirow{5}{*}{\textbf{DATA.}}} & \multicolumn{1}{l|}{CitySim~\cite{robinson2009citysim} / NGSIM~\cite{usdot2016ngsim}} & 
\no & \no & \no & \yes & \no & \no & \no & \no \\

\multicolumn{1}{c|}{} & \multicolumn{1}{l|}{Bench2Drive~\cite{jia2024bench2drive}} & 
\no  & \no & \no & \no & \yes & \yes & \no & \no \\

\multicolumn{1}{c|}{} & \multicolumn{1}{l|}{DriveLM-CARLA~\cite{sima2023drivelm}} & 
\no  & \no & \no & \no & \yes & \yes & \no & \no \\

\multicolumn{1}{c|}{}& \multicolumn{1}{l|}{nuPlan dataset~\cite{caesar2021nuplan}} & 
\no & \no & \yes & \yes & \no & \yes & \no & \no \\

\multicolumn{1}{c|}{}& \multicolumn{1}{l|}{nuScenes~\cite{caesar2020nuscenes} / Waymo dataset~\cite{sun2020scalability}} & 
\no & \no & \yes & \yes & \yes & \yes & \no & \no \\

\midrule

\multicolumn{1}{c|}{\multirow{2}{*}{\textbf{GEN.}}} & \multicolumn{1}{l|}{MagicDrive~\cite{gao2023magicdrive} / DriveDreamer~\cite{wang2023drivedreamer}} & 
\no & \no & \yes & \yes & \yes & \yes & \no & \no \\

\multicolumn{1}{c|}{}& \multicolumn{1}{l|}{SimGen~\cite{zhou2024simgen}} & 
\no & \no & \yes & \yes & \yes & \no & \no & \no \\ \midrule

\multicolumn{1}{c|}{\multirow{3}{*}{\textbf{W.M.}}} & \multicolumn{1}{l|}{KiGRAS~\cite{zhao2024kigras} / SMART~\cite{wu2024smart}} & 
\yes & \no & \no & \yes & \no & \no & \no & \yes \\

\multicolumn{1}{c|}{}& \multicolumn{1}{l|}{MUVO~\cite{bogdoll2023muvo}} & 
\yes & \no & \no & \yes & \no & \no & \no & \no \\

\multicolumn{1}{c|}{}& \multicolumn{1}{l|}{Vista~\cite{gao2024vista} / GAIA-1~\cite{hu2023gaia}} & 
\yes & \no & \yes & \no & \yes & \no & \no & \no \\ \midrule

\multicolumn{1}{c|}{\multirow{5}{*}{\textbf{SIM.}}} & \multicolumn{1}{l|}{Waymax~\cite{gulino2024waymax}} & 
\yes & \yes & \no & \yes & \no & \no & \no & \no \\

\multicolumn{1}{c|}{} & \multicolumn{1}{l|}{SUMO~\cite{krajzewicz2012recent} / LimSim~\cite{wenl2023limsim}} & 
\yes & \yes & \no & \yes & \no & \no & \no & \yes \\

\multicolumn{1}{c|}{} & \multicolumn{1}{l|}{CARLA~\cite{dosovitskiy2017carla}} & 
\yes & \yes & \no & \yes & \yes & \yes & \yes & \yes \\

\multicolumn{1}{c|}{} & \multicolumn{1}{l|}{MetaDrive~\cite{li2022metadrive}} & 
\yes & \yes & \no & \yes & \no & \yes & \yes & \yes \\

\multicolumn{1}{c|}{} & \multicolumn{1}{l|}{Unisim~\cite{yang2023unisim} / OAsim~\cite{yan2024oasim}} & 
\yes & \yes & \yes & \yes & \no & \yes & \no & \no \\

\midrule

\multicolumn{1}{c|}{\textbf{Ours}} & \multicolumn{1}{l|}{\textbf{\arena}} & \yes & \yes & \yes & \yes & \yes & \yes & \yes & \yes \\ \bottomrule
\end{tabular}
}
\caption{Comparison of various datasets, generative models, world models, and simulators in terms of interactivity, fidelity, and diversity features. \textbf{DATA.} represents dataset, \textbf{GEN.} represents generative model, \textbf{W.M.} represents world model, \textbf{SIM.} represents simulator. }
\label{tab:comparison}
\end{table*}

\subsection{Open-loop and Closed-loop Experiments}

In this section, we adopt the prevailing end-to-end autonomous driving method UniAD~\cite{hu2023planning} as the driving agent to test both the open-loop and closed-loop performance within the {\arena} framework. We utilized UniAD's open-source code and pre-trained weights without additional training. UniAD operates at 2Hz, outputting a trajectory of 6 path points over the next 3 seconds. {\engine} further interpolates this to a 10Hz trajectory.

\noindent \textbf{Open-loop Evaluation.}
We first assess UniAD's performance in {\arena}'s open-loop mode.
UniAD is evaluated on three scenarios: \textit{1)} the original nuScenes image sequences; \textit{2)} {\dreamer}-generated nuScenes image sequences, where the vehicles' trajectory remains identical to nuScenes ground truth, but surround images are replaced with {\dreamer}-generated ones; and \textit{3)} \arena's own simulation sequences (i.e., {\arena}'s open-loop mode).
Our evaluation metrics consist of the PDM Scores and its sub-scores, as detailed in Section~\ref{sec: simu modes}. Additionally, we evaluate trajectories driven by human drivers in nuScenes as the human driver performance. Detailed results are presented in Table~\ref{tab:open-loop perform}.

The results reveal that while UniAD performs optimally on the original nuScenes sequence with a PDMS metric of 0.91, the {\dreamer}-generated sequence surprisingly achieves a PDMS of 0.902, representing a metric drop of less than 1\%. We attribute this to both the high fidelity of our {\dreamer}-generated images and UniAD's strong dependence on ego states, as corroborated by~\cite{li2024ego}.

In {\arena}'s open-loop mode, Figure~\ref{fig:uniad openloop} illustrates two sequences, demonstrating that UniAD's prediction of the road network and vehicle tracking are fundamentally accurate. However, in terms of metrics, UniAD's performance in such scenarios with unseen road and traffic flow is significantly degraded, with an average PDM Score of only 0.636. The output trajectories by UniAD exhibit a substantial increase in collision rates and instances of driving outside the drivable area.
The open-loop experimental results underscore the critical importance of closed-loop experiments and tests for autonomous driving methods.

\noindent \textbf{Closed-loop Evaluation.}
We further evaluated UniAD's performance in {\arena}'s closed-loop mode. In this mode, the trajectory outputted by UniAD is directly used for ego vehicle control, and the evaluation metrics include PDM Score (PDMS), Route Completion (RC), and Arena Drive Score (ADS). Our closed-loop experiment was conducted on four pre-set paths, with two paths selected in Boston and two in Singapore. The simulation time to complete each trajectory was approximately 120 seconds. Detailed results are presented in Table~\ref{tab:uniad closed-loop perform}.

The results indicate that the PDMS of UniAD-generated trajectories in closed-loop mode (0.667) is comparable to that of the open-loop mode. However, the Route Completions (RC) are consistently low, averaging only about 13.7\% of the total route length. Specifically, UniAD performs better on straightaways but largely fails to navigate the first turning intersection in the route. Figure~\ref{fig:uniad fail1} illustrates two failure cases where UniAD lacked sufficient trajectory correction. Despite a roughly correct prediction of the road structure, it ultimately mounted the central green belt or failed to complete a right turn successfully. The average Arena Driving Score for UniAD is 0.086.
It should be noted that these are preliminary results based on testing only 4 routes. We plan to expand the number of routes for a more comprehensive evaluation and explore the combined effect of {\dreamer}'s timing consistency and the driver agent's performance on the final ADS.

\section{Related Works}
\label{sec:related_work}

\subsection{Data Acquisition for Autonomous driving}
\label{sec:data source}
The characteristics of the automated driving dataset can be categorized into two aspects: appearance fidelity and interactivity. First, in terms of appearance fidelity, NGSIM~\cite{usdot2016ngsim} and CitySim~\cite{robinson2009citysim} provide only realistic driving trajectories that can provide safe and reliable driving planning guidance. On top of that, some datasets developed based on the Carla simulator, such as DriveLM-CARLA~\cite{sima2023drivelm} and BenchDrive~\cite{jia2024bench2drive}, provide simulated sensor data. Taking it a step further, the Waymo~\cite{sun2020scalability} and nuScenes~\cite{caesar2020nuscenes} datasets capture real-world sensor recordings and the driving behavior of human drivers. The datasets were produced in a complex process and with a limited amount of data. To add variety to the scenarios, MagicDrive~\cite{gao2023magicdrive} and DriveDreamer~\cite{wang2023drivedreamer} provide editable scenario generation. So far, we have been able to obtain diverse and rich data for training. However, the above data can only be used for open-loop evaluation, i.e., current decisions do not affect future data distributions, which differs significantly from real driving. Works~\cite{hu2023gaia,gao2024vista,bogdoll2023muvo,zhao2024kigras,wu2024smart} that also have fidelity differences, improve the interactivity of the data, they usually use auto-regressive generation methods to realize the interaction, the generation process implies the model's understanding of the world, and usually can not be too much human intervention. Some simulators~\cite{gulino2024waymax,wenl2023limsim,li2022metadrive,dosovitskiy2017carla,yang2023unisim,yan2024oasim,krajzewicz2012recent} make things more controllable by decoupling part of the mechanics of how the world works. Common examples include simulators~\cite{krajzewicz2012recent,gulino2024waymax,wenl2023limsim} that provide realistic traffic flow, and simulators~\cite{dosovitskiy2017carla, li2022metadrive} that drive vehicles in game engines, and reconstructive simulations represented by~\cite{yang2023unisim,yan2024oasim} that provide the appearance of reality.

\subsection{Diffusion-based Generative Models}
Recent advancements in generative models have seen diffusion models play a pivotal role in image and video generation~\cite{dhariwal2021diffusion, meng2021sdedit, nichol2021glide, podell2023sdxl, ramesh2022hierarchical, blattmann2023stable, he2022latent}. Moreover, recent works have expanded the scope by integrating additional control signals beyond traditional text prompts~\cite{guo2023animatediff, li2023gligen, mou2024t2i}. For instance, ControlNet~\cite{zhang2023adding} incorporates a trainable version of the SD encoder for control signals, while studies such as Uni-ControlNet~\cite{zhao2024uni} and UniControl~\cite{qin2023unicontrol} have emphasized the fusion of multi-modal inputs into a unified control condition using input-level adapter structures. 
Our approach aims to study the generation of continuous and controllable sequence frames, thereby bridging the gap between simulation environments and reality, and establishing the required foundation for closed-loop learning of autonomous driving agents.

\subsection{Evolution of Autonomous Driving Generation} 

World Models~\cite{hu2023gaia, yang2024generalized} utilize diffusion models to generate future driving scenes based on historical information, these methods often lack the ability to control the scenarios through layout, are difficult to generate continuous and stable videos and lack the approximation of physical laws.
TrackDiffusion focused on generating videos based on 2D object layouts~\cite{li2023trackdiffusion}.
BEVGen~\cite{swerdlow2024street} pioneered the generation of synthetic multi-view images based on the BEV layout, laying the foundation for a controllable generation of autonomous driving scenarios. BEVControl~\cite{yang2023bevcontrol} extended this approach by a height elevation process, enabling image generation aligned with surrounding projection layouts.
Further advancements includes MagicDrive~\cite{gao2023magicdrive}, DriveDreamer~\cite{wang2023drivedreamer}, Panacea~\cite{wen2024panacea} and DrivingDiffusion~\cite{li2023drivingdiffusion}, which generate panoramic controllable videos through various 3D controls and encoding strategies.
However, their primary focus lies in augmenting training data to enhance algorithm performance, rather than serving as simulators for modeling dynamic environmental interactions.

\subsection{Simulator-Driven Scenario Generation}
Autonomous vehicle development is significantly enhanced by driving simulators, which provide controlled environments for realistic simulation. Prominent research efforts have concentrated on generating virtual imagery and annotations, with some studies expanding to incorporate environmental variations and construct safety-critical scenarios for training based on real-world data logs. Nevertheless, these simulated images frequently fall short of achieving true realism, as evidenced by previous works~\cite{ros2016synthia, richter2016playing, sun2022shift}. 
While SimGen~\cite{zhou2024simgen} made a breakthrough as the first work to generate diverse driving scenarios following conditions from a simulated environment, it mainly focused on the quality of the generated content with only front-view images, neglecting the exploration of closed-loop systems. Our research aims to bridge this gap by developing a system that can not only generate realistic scenarios but also allow agents to interact with them in a closed-loop manner.

\subsection{Closed-Loop Driving in Simulation}
End-to-end vehicle control algorithms~\cite{hu2022st, hu2023planning, ye2023fusionad}, are typically trained and evaluated on open-loop datasets~\cite{caesar2020nuscenes}. However, these algorithms lack the capability to generalize directly to simulators for closed-loop evaluation, which hinders the demonstration of their true performance potential. Recent studies have increasingly recognized the significance of closed-loop evaluation, as exemplified by~\cite{jiang2023vad, wang2023drivemlm}. Moreover, simulation environments offer a wealth of training data, a stark contrast to models trained on datasets that are constrained by data distribution~\cite{li2024ego}. A significant challenge arises due to the discrepancy between the simulated scene's appearance and real-world conditions, complicating the generalization of models trained on simulation data to actual scenarios. This creates a paradox: the desire to utilize simulation data for its diversity and editability, while also seeking data that closely mirrors reality. Our approach effectively addresses this issue by enhancing the realism of the simulator for certain closed-loop learning methods~\cite{mei2024continuously}.

\section{Conclusions and Future Works}
This paper introduces a novel closed-loop simulation platform named {\arena} for vision-based driving agents. {\arena} integrates a {\engine} that generates human-like traffic flow and a high-fidelity generative {\dreamer} with infinite generation. This combination allows realistic interaction and continuous feedback between the driving agent and the simulation environment. The system provides a valuable platform for developing and testing autonomous driving agents in a variety of scenarios, marking a substantial leap in driving simulation technology.

{\arena} is designed with a modular architecture, allowing for easy replacement of each module. This paper presents an initial implementation of {\arena}. As the first high-fidelity closed-loop simulator, we still have a few limitations for future improvement:

1) Data Diversity: The current generative model is trained solely on the nuScenes dataset, which limits the diversity and emergence capabilities. We plan to expand training to include more varied datasets to enhance the model's robustness and versatility.

2) Temporal Consistency: While we can generate continuous videos with an autoregression strategy, maintaining motion trends and temporal consistency between frames remains challenging. Future work will explore multi-frame autoregressive networks and more scalable architectures~\cite{peebles2023scalable} to address these issues.

3) Runtime Efficiency: Like many generative models, {\dreamer} requires significant runtime. Investigating faster sampling methods~\cite{lu2022dpm} and model quantization may alleviate these problems.

4) Expanded Agent Testing: We plan to incorporate a broader range of driving agents within {\arena}, facilitating the continuous learning and evolution of knowledge-driven driving agents in the closed-loop environment~\cite{li2023towards}.

5) A Real Arena: {\arena} can not only evaluate the performance of different driving agents, but also act as a testing ground for AD generative models. By using the same driving agent as a referee, it can fairly assess the sim-to-real gap of different generative models. This approach even provides a more credible and convincing evaluation compared to traditional metrics like FID and FVD.

We recognize that practical application may still be a way off, but the potential and promise shown by this work are evident. We hope this research will advance closed-loop exploration in highly realistic environments and offer a valuable platform for developing and assessing driving agents across a range of challenging scenarios. We encourage the community to collaborate in advancing this field. The era of open loops is transitioning, and autonomous driving evaluation and learning are set to enter a new era of closed-loop systems.

\section*{Acknowledgments}
The research was supported by Shanghai Artificial Intelligence Laboratory, the National Key R\&D Program of China (Grant No. 2022ZD0160104) and the Science and Technology Commission of Shanghai Municipality (Grant Nos. 22DZ1100102 and 23YF1462900).

\clearpage
\bibliographystyle{ieeetr}
\bibliography{egbib}
\end{document}